\documentclass[conference]{IEEEtran}
\IEEEoverridecommandlockouts

\usepackage{cite}
\usepackage{amsmath,amssymb,amsfonts}
\usepackage{algorithmic}
\usepackage{algorithm}
\usepackage{graphicx}
\usepackage{textcomp}
\usepackage{xcolor}
\def\BibTeX{{\rm B\kern-.05em{\sc i\kern-.025em b}\kern-.08em
    T\kern-.1667em\lower.7ex\hbox{E}\kern-.125emX}}
\begin{document}

\title{KAN v.s. MLP for Offline Reinforcement Learning
\thanks{*Corresponding Authors. Emails: li.jiao@imicams.ac.cn; \\
liuhy@sem.tsinghua.edu.cn}
}
\author{Haihong Guo$^{1,2,3}$, Fengxin Li$^{1,3}$, Jiao Li$^{2,*}$, Hongyan Liu$^{4,*}$\\
\\
$^{1}$ School of Information, Renmin University of China, China\\
$^{2}$ Institute of Medical Information / Medical Library, Chinese Academy of Medical Sciences / \\
Peking Union Medical College, China\\
$^{3}$ Key Laboratory of Data Engineering and Knowledge Engineering, Ministry of Education, China\\
$^{4}$ School of Economics and Management, Tsinghua University, China}

\maketitle

\begin{abstract}
Kolmogorov-Arnold Networks (KAN) is an emerging neural network architecture in machine learning. It has greatly interested the research community about whether KAN can be a promising alternative of the commonly used Multi-Layer Perceptions (MLP). Experiments in various fields demonstrated that KAN-based machine learning can achieve comparable if not better performance than MLP-based methods, but with much smaller parameter scales and are more explainable. In this paper, we explore the incorporation of KAN into the actor and critic networks for offline reinforcement learning (RL). We evaluated the performance, parameter scales, and training efficiency of various KAN and MLP based conservative Q-learning (CQL) on the the classical D4RL benchmark for offline RL. Our study demonstrates that KAN can achieve performance close to the commonly used MLP with significantly fewer parameters. This provides us an option to choose the base networks according to the requirements of the offline RL tasks.
\end{abstract}

\begin{IEEEkeywords}
Kolmogorov-Arnold networks, KANs, multi-layer perceptrons, MLPs, offline reinforcement learning
\end{IEEEkeywords}

\begin{figure*}[h]
\centerline{\includegraphics[width=0.9\textwidth]{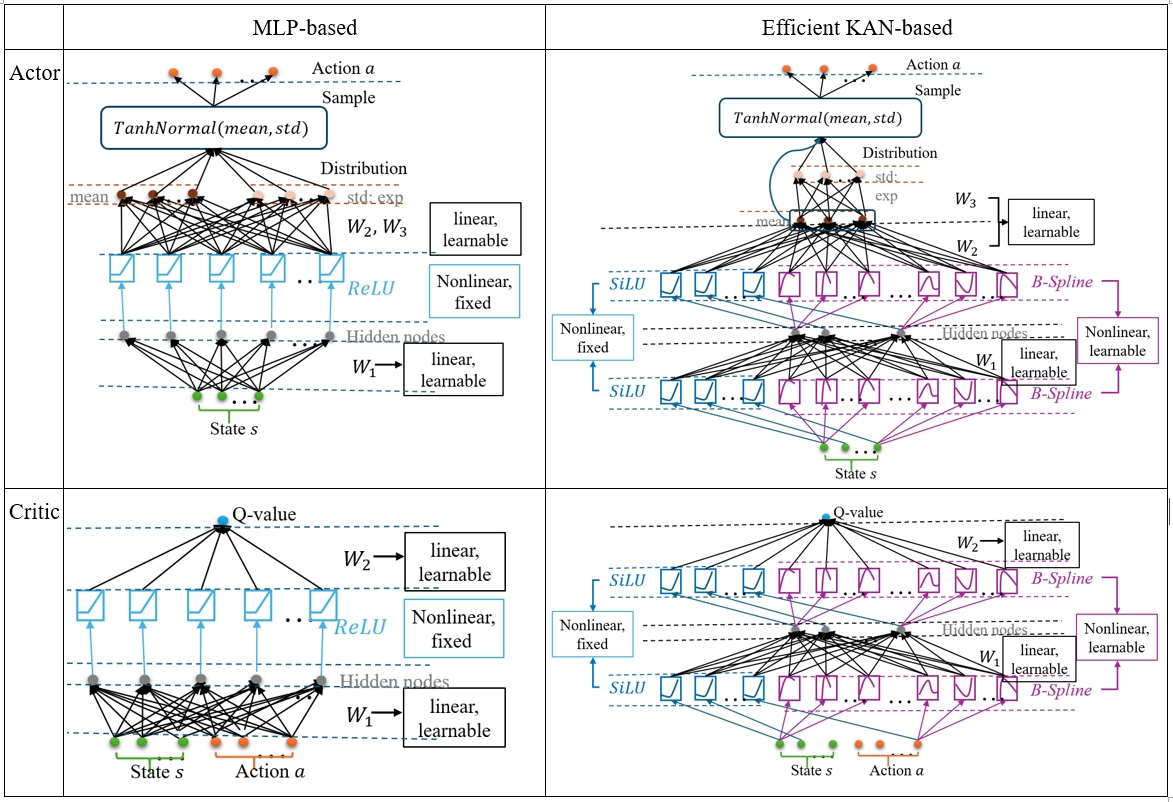}}
\caption{The overall architecture of RCSD. }
\label{fig: CQL_KAN_MLP}
\end{figure*}

\section{Introduction}
Kolmogorov-Arnold Networks (KAN) \cite{KAN} is an emerging neural network architecture in machine learning, attracting widespread attention ever since it was hang on arXiv on April 30, 2024. The research community is very concerned about whether KAN can be a promising alternative of the commonly used Multi-Layer Perceptions (MLP). 

KAN and MLP are different in the fundamental theorems and network architectures, and KAN is thought to enjoy super flexibility and interpretability than MLP. Empirical studies demonstrated that KAN excels in data fitting and partial differential equation solving \cite{KAN_time_series, KAN_PDE, KAN_PDE2}, high-dimensional data handling \cite{KAN_time_series2}, graph-structured data processing and hyperspectral image classification \cite{KAN_image}, complex data relationships capturing \cite{KAN_relationship}, and online reinforcement learning (RL) \cite{KAN_RL}, etc. showing that even smaller KAN-based models can achieve or surpass the performance of MLP-based models in these tasks. The efficacy of KAN for offline RL has not been studied yet. Therefore, this paper aimed at applying KAN as an alternative to MLP in offline RL and comparing their performance, parameter scales and training efficiency.

Offline RL leverages previously collected offline datasets without further direct environment interaction to optimize sequential decision policy \cite{Levine2020Review_ORL}. Offline RL has the capacity to provide a safer and more economically viable solution for real-world sequential decision-making comparing to online RL  \cite{Agarwal2020Perspective_ORL, Prudencio2023Review}. The main challenge of offline RL is to avoid extrapolation errors caused by distribution shift when learning policies with limited offline datasets. Conservative Q-learning (CQL) \cite{Kumar2020CQL} is a state-of-the-art offline RL method which can alleviate extrapolation error through conservatively estimating the value of out-of-distribution actions. CQL is originally based on MLP.

Networks with fewer parameters yet equal or superior approximation capabilities have the potential to significantly enhance performance in various offline RL applications. In this work, we propose the use of KAN as the basic building blocks of the actor and critic networks for CQL. The main contributions of this paper are summarised as following: 1) The first application of KAN in a offline RL algorithm. 2) Comprehensive comparison of performance, parameter scales, and training efficiency between KAN-based, MLP-based, and KAN-MLP-based CQL in various continuous control tasks.

\section{RELATED WORKS}

KAN has recently attracted significant attention due to its versatility and enhanced application performance \cite{Temporal_KAN, EHD_KAN, KAN_time_series3, KAN_time_series4, KAN_evolutionary, DeepOKAN, KAN_feature_extractors, KAN_tabular, KAN_Vision, CEST-KAN}. Experiments in different fields have shown that KAN-based machine learning can achieve comparable, if not better, performance than MLP-based methods while using much smaller parameter scales and offering better explainability \cite{KAN_review}. For example, Liu et al. \cite{KAN_time_series,KAN2.0} demonstrated KAN’s superior performance in symbolic regression, effectively modeling complex data and achieving precise formula discovery. Recent studies integrated path signatures with KAN to better understand and predict complex time series patterns \cite{KAN_path_signature,KAN_path_signature2}. 

GKAN\cite{GKAN} and GraphKAN \cite{GraphKAN} extends their principles to graph-structured data through node feature aggregation, embedding layer processing, and dynamic adaptability. Wav-KAN\cite{Wav-KAN} introduces wavelet functions as learnable activation functions into the KAN architecture to achieve nonlinear mapping of input spectral features for hyperspectral image classification. SpectralKAN \cite{SpectralKAN} uses B-spline functions as activation functions to represent multivariate continuous functions for feature extraction and classification of hyperspectral images. The VQ-KAN\cite{VQ-KAN} model is specifically designed for quantum computing optimization problems, utilizing KAN’s flexibility by employing learnable univariate activation functions to capture and optimize the complex relationships between quantum states and operations.  U-KAN\cite{U-KAN} integrates KAN layers into the U-Net architecture, enhancing the accuracy and interpretability of medical image segmentation models. Some studies \cite{KAN_robust, KAN_robust2} explored and improved the robustness of KAN in adversarial scenarios, highlighting the potential of KAN as a secure and reliable model in environments where robustness is crucial. To the best of our knowledge, there have been no efforts to apply KAN as an alternative to MLP in offline RL and compare their performance, parameter scales, and training efficiency, making this study a pivotal contribution.

\begin{table*}[h]
\caption{Various configurations of KAN and MLP based CQL}
\centering
\begin{tabular}{c  |ccc  ccc  c  c  c  c}
\hline
Name&\multicolumn{3}{c}{\textbf{MLP-based CQL}}&\multicolumn{3}{c}{\textbf{KAN-based CQL} }& \multicolumn{4}{c}{\textbf{KAN-MLP-based CQL}}\\
 & \textbf{a1c1}&  \textbf{a2c2}&\textbf{a3c3}& \textbf{a0c0}& \textbf{a1c1}& \textbf{a2c2}& \textbf{a0c3}& \textbf{a1c3}& \textbf{a2c3} &\textbf{a3c3}\\ \hline
Actor network&MLP& MLP&MLP&KAN&  KAN&KAN& KAN& KAN& KAN&KAN\\
Actor hidden layers&1& 2&3&0&  1&2& 0& 1& 2&3\\
 Actor hidden size&256&  256&256& 64&  64&64& 64& 64& 64&64\\
 Critic Networks& MLP& MLP& MLP& KAN& KAN& KAN& MLP& MLP& MLP&MLP\\
 Critic hidden layers& 1& 2& 3& 0& 1& 2
& 3& 3& 3&3\\ \hline
Critic hidden size&256& 256&256&64&  64&64& 256& 256& 256&256\\ \hline
\end{tabular}
\end{table*}

\begin{table*}[h]
\caption{Normalized scores on 9 continuous tasks in the D4RL benchmark. ME, MR, and M represent medium-expert, medium-replay, and medium datasets respectively. The top 2 best results are bold.  }
\centering
\begin{tabular}{c  |llc  llc  c  c  c  l}
\hline
\textbf{Task} &\multicolumn{3}{c}{\textbf{MLP-based CQL}}&\multicolumn{3}{c}{\textbf{KAN-based CQL} }&\multicolumn{4}{c}{\textbf{KAN-MLP-based CQL}}\\
 & \textbf{a1c1}&  \textbf{a2c2}&\textbf{a3c3}& \textbf{a0c0}& \textbf{a1c1}& \textbf{a2c2}&\textbf{a0c3}& \textbf{a1c3}& \textbf{a2c3} & \textbf{a3c3}\\ \hline
Walker2d-ME&-0.2±0.0& 101.8±20.0&\textbf{109.5±0.5}&3.4±0.0&  31.3±30.0&93.7±20.0&0.8±0.0& 105.2±10.0& \textbf{107.7±0.0} &  105.5±10.0
\\
Hopper-ME&0.7±0.0& 17.7±10.0&\textbf{94.8±24.7}&0.6±0.0&  91.3±30.0&76.2±30.0&12.3±0.0& 58.4±30.0& 93.4±20.0& \textbf{95.4±20.0}\\
HalfCheetah-ME&2.9±0.0& 25.4±20.0&\textbf{79.1±11.6}&-0.7±0.0&  1.2±0.0&12.0±10.0&4.8±0.0& 49.1±20.0& \textbf{55.6±20.0} & 54.1±20.0
\\ \hline
Walker2d-MR&& 80.3±10.0
&\textbf{84.1±9.4}&-0.6±0.0
&  28.7±20.0&51.1±30.0&2.5±0.0& 78.8±10.0& \textbf{84.9±10.0} & 82.2±10.0
\\
Hopper-MR &1.8±0.0& 67.9±20.0
&\textbf{97.4±3.2}&0.7±0.0
&  53.8±20.0&75.4±30.0&5.4±0.0& 90.1±20.0& \textbf{94.4±10.0} & 93.9±10.0
\\
 HalfCheetah-MR&11.4±10.0&  \textbf{45.0±0.0}&\textbf{44.4±0.8}& -1.3±0.0
&  34.5±0.0&40.2±0.0&8.0±0.0& 41.6±0.0& 43.7±0.0
 & 43.9±0.0
\\ \hline
Walker2d-M&-0.2±0.0& 79.7±0.0&\textbf{82.7±1.3}&-0.4±0.0&  20.1±30.0&81.0±10.0&6.2±10.0& \textbf{82.1±0.0}& 81.4±0.0 & 79.4±10.0
\\
Hopper-M&0.6±0.0& \textbf{66.4±10.0}&\textbf{65.1±14.0}&0.6±0.0&  54.1±10.0&53.2±10.0&19.5±10.0& 54.8±10.0& 56.4±10.0
 & 61.1±10.0
\\
 HalfCheetah-M&-1.3±0.0&  44.9±0.0&\textbf{46.3±0.7}& -0.7±0.0&  36.7±10.0&42.5±0.0&7.5±0.0& 43.7±0.0& 44.5±0.0& \textbf{45.4±0.0}\\ \hline
Average &2.0& 58.8 
&\textbf{78.2}&0.2 
&  39.1 &58.4 &7.4 & 67.1 & \textbf{73.6} & 73.4 
\\ \hline
\end{tabular}
\end{table*}

\begin{table*}[h]
\caption{Number of parameters in actor networks and average training time per epoch (unit: s)}
\centering
\begin{tabular}{c  |llc  llc  c  c  c  l}
\hline
\textbf{Environment}&\multicolumn{3}{c}{\textbf{MLP-based CQL}}&\multicolumn{3}{c}{\textbf{KAN-based CQL} }& \multicolumn{4}{c}{\textbf{KAN-MLP-based CQL}}\\
 & \textbf{a1c1}&  \textbf{a2c2}&\textbf{a3c3}& \textbf{a0c0}& \textbf{a1c1}& \textbf{a2c2}& \textbf{a0c3}& \textbf{a1c3}& \textbf{a2c3} &\textbf{a3c3}\\ \hline
Walker2d&7,692& 73,484&139,276&1,062&  14,762&55,722& 1,062& 14,762& 55,722
 &96682 
\\
Hopper&4,614& 70,406&136,198&342 &  8,972&49,932& 342 & 8,972& 49,932
 &90892 
\\
 HalfCheetah&7,692&  73,484&139,276& 1,062&  14,762&55,722& 1,062& 14,762& 55,722
 &96682 
\\ \hline
Training time&17.3 & 20.4 &21.4 &31.3 &  58.3 &87.0 & 27.4 & 42.8 & 56.0 
 &59.2\\ \hline
\end{tabular}
\end{table*}

\begin{figure}[h]
\centering
\includegraphics[width=0.5\textwidth]{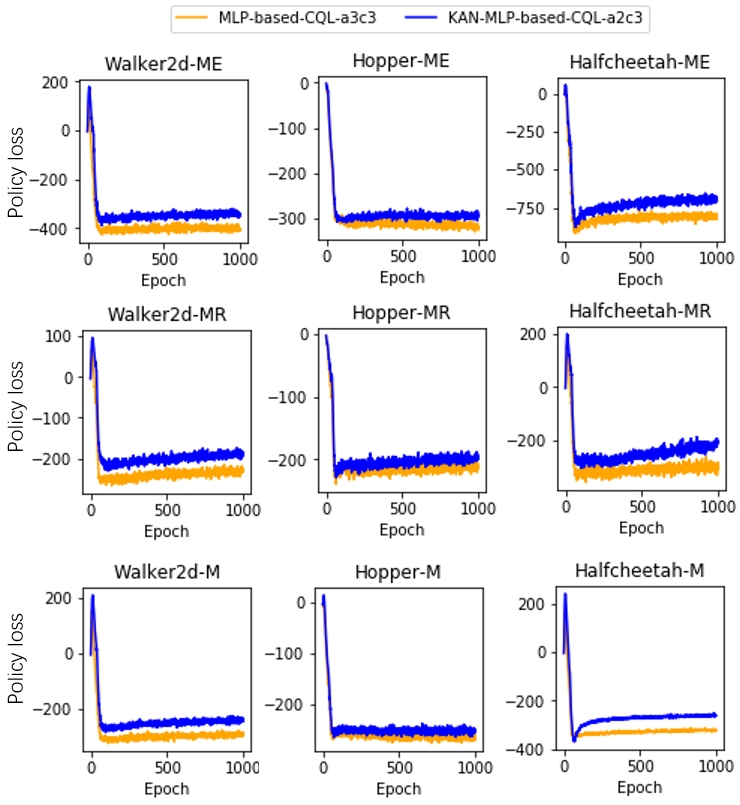}
\caption{Comparison of policy loss between MLP-based and KAN-MLP-based CQL}
\label{fig: Performance comparison}
\end{figure}

\section{BACKGROUND}
\textbf{Offline Reinforcement Learning (RL)} RL is developed for sequential decision-making, involving an agent and its environment. The agent follows a policy  \(\pi(\cdot|s)\), making decisions and taking action $a$ at each time step according to its current state $s$, resulting in a transition to the next state $s^\prime$ and receiving a numerical reward $r$ from the environment \cite{Sutton2018RL-book}. The policy is optimized by selecting the action with the maximum value, i.e. the maximum cumulative return. 

In the online setting, the agent can learn the optimal policy through trail and error by continuously interaction with the environment. However, in the offline setting, the environment is inaccessible.  The agent learns policy with previously collected static dataset \(\mathcal{D}=\left\{\left(s,\ a,\ r,\ s^\prime,\ done\right)\right\}\). Thus the main challenge is to avoid extrapolation errors caused by distribution shift when learning policies with limited offline datasets. 

\textbf{MLP-based Conservative Q-learning (CQL)} CQL\cite{Kumar2020CQL} is a state-of-the-art offline RL algorithm developed upon soft actor-critic (SAC) \cite{Haarnoja2018SAC}.  It uses an actor network $\pi(\cdot|s)$ as the policy function to generate action, and two critic networks \(Q_1(s,a)\) and \(Q_2(s,a)\) to estimate the action value, also referred to as Q-value. CQL seeks to avoid extrapolation error through penalizing the Q-value of OOD actions while rewarding that of in-distribution actions \cite{Kumar2020CQL}. The loss functions of the actor and critic networks are as following:
\begin{align}\label{eq: CQL}
\mathcal{L}_{c_i}=\alpha_1(\mathbb{E}_{s\sim\mathcal{D},a\sim\pi(a|s)}\left[Q_{i}(s,a)\right]-\mathbb{E}_{s,a\sim\mathcal{D}}\left[Q_i(s,a)\right]) \\ \notag
+\frac{1} {2}\mathbb{E}_{s,a,s^\prime\sim\mathcal{D}}\left[\left(Q_i(s,a)-\hat Q(s,a)\right)^2\right] + \mathcal{R}(\pi)
\end{align}

\begin{align}\label{eq: policy smooth loss}
\mathcal{L}_{a}=\mathbb{E}_{s\in\mathcal{D},a\sim\pi(\cdot|s)}&\left[-\min_{j=1,2}{Q_{j}\left(s,a\right)}+\alpha_2\cdot l o g\pi\left(a|s\right)\right]
\end{align}

where $\hat{Q}(s,a)$ is the soft temporal difference (TD) target Q-value, computed as
$r + \gamma \mathbb{E}_{a^\prime \sim \pi(s^\prime)} \left[\min_{j=1,2} Q_j^\prime(s^\prime, a^\prime)  - \alpha_2\cdot \log \pi(a^\prime | s^\prime) \right]$  \cite{Haarnoja2018SAC}, $Q_j^\prime$ is a target Q-network, $\mathcal{R}(\pi_\theta)$ is a regularizer which is chosen to be the KL-divergence against a prior distribution Unif(a), and $\alpha_1$ and $\alpha_2$ are two  automatic turned hyper parameters.

Originally, the actor and critic networks in CQL are constructed based on MLP. The structure diagrams with one hidden layer are shown in the left in Fig. \ref{fig: CQL_KAN_MLP}. In MLP-based CQL, each layer of the actor and critic networks first go through a linear transformation $X=W_iX+b$, then a fixed nonlinear transformation, such as ReLU($X$), where $W_i, i=1,2,...$, are learnable weight parameters, and $b$ is the bias. The policy is designed as TanhGaussianPolicy \cite{Haarnoja2018SAC}, so when we get the mean and standard deviation by the linear layer, we need to convert them into a TanhNormal distribution, and then  sample from the distribution to get stochastic continuous actions.

\textbf{Kolmogorov-Arnold Networks (KANs)} KAN is a new building block in machine learning which is regarded as a promising alternative to MLP. The main differences between KAN and MLP are \cite{KAN, KAN_review}: 1) \textbf{The fundamental theorems behind their design}: MLP is based on the Universal Approximation Theorem \cite{UAT}, which states that a feed forward network with enough hidden nodes and a suitable non-linear activation function can approximate any continuous function. On the other hand, KAN is inspired by the Kolmogorov-Arnold Theorem \cite{KAN_theorem1,KAN_theorem2}, which shows that any continuous multi-variable function can be broken down into a sum of a finite number of non-linear functions. 2) \textbf{Representation of weight parameters}: MLP represents weight parameters using learnable full-connected linear functions, while KAN uses learnable univariate non-linear functions for each weight parameter. 3) \textbf{Nonlinear activation functions}: MLP uses predefined fixed nonlinear activation functions (such as sigmoid, ReLU, etc.), while KAN uses learnable activation functions (such as B-Spline) that can be adjusted during training to adapt to complex data patterns and nonlinear relationships. 4) \textbf{Flexibility in architecture}: The architecture of MLP is fixed and cannot easily incorporate domain knowledge, while the architecture of KAN can be customized and adjusted according to specific tasks and datasets, allowing for the incorporation of domain knowledge into the univariate functions \cite{KAN_knowledge}. The original implementation of KAN had performance bottlenecks limiting its use in large-scale problems. There are various improvements\cite{fKAN, BSRBF-KAN, convolutional_KAN, rKAN}, and efficient KAN \cite{Efficient_KAN} has been developed to improve computing efficiency and performance.

\section{METHOD}
 
\subsection{Network Architecture}\label{AA}
We use CQL as the core framework of offline RL and incorporate efficient KAN into its actor and critic networks respectively. The structure diagrams with one hidden layer KAN are shown in the right in Fig. \ref{fig: CQL_KAN_MLP}. 

Overall, the KAN-based actor network and critic networks have the same middle layers, which is defined as $H_{i}=W_i\cdot [\text{SiLU}(H_{i-1}),\text{B-Spline}(H_{i-1})]$, where SiLU is a fixed nonlinear activation function and B-Spline is a learnable nonlinear function, and $H_i,i=1,2,...$ are hidden or output layers. One of the difference is that the input of the actor network is state, i.e. $H_1=W_1\cdot [\text{SiLU}(s),\text{B-Spline}(s)]$, while the input the critic networks are (state, action) pairs, i.e. $H_1=W_1\cdot [\text{SiLU}(s,a),\text{B-Spline}(s,a)]$. Another difference is that the output layer of the critic networks has only one node, representing the Q-value of the  (state, action) pairs. The actor network needs to get the get the mean of the TanhGaussianPolicy, then conduct a linear transformation to get the standard deviation. The remaining processes are the same as the MLP-based actor network.

We designed compared various KAN and MLP based CQL architectures. The configurations are shown in Table 1. Other parameters are kept the same as the original MLP-based CQL and the efficient KAN.  

\subsection{BENCHMARK ENVIRONMENT AND DATASETS}\label{AA}
To evaluate the performance of various KAN and MLP based CQL, we conduct experiments on the classical D4RL benchmark \cite{Fu2020D4RL} for offline RL. We evaluate each method on 9 continuous Gym tasks, which involves three different environments (Walker2d, Hopper, and HalfCheetah) with three levels of dataset (medium-expert, medium-relay, and medium) for each environment.  We will release the code once the paper is accepted. 

\section{RESULTS}

\subsection{Performance comparison  }\label{AA}
Table 2 reports the performance of each method, measured by normalized scores with standard deviation. The score is the return gained at the end of each episode and is normalized to D4RL scores that measure how the learned policy compared with an expert policy and a random policy that are pre-installed in D4RL: $\text{normalized score} = 100 \times\frac{(\text{learned policy score}-\text{random policy score})}{(\text{expert policy score}-\text{random policy score})}$. The results are averaged over 4 random seeds. We have the following observations: 1) The MLP-based CQL with 3 hidden layers both for the actor and the critic networks enjoys the best performance; 2) The KAN-MLP-based CQL with 2 hidden layers in the KAN-based actor and 3 hidden layers in the MLP-based critic performed close to the best MLP-based CQL, but the latter is slightly better in most of tasks (Fig \ref{fig: Performance comparison}), and increase hidden layers in the KAN-based actor can not gain performance improvements. 3) Pure KAN-based CQL is inferior to MLP-based CQL and KAN-MLP-based CQL when the nonlinear transformation layers are the same. 

\subsection{Parameter scale and training efficiency comparison }\label{AA}
Table 3 shows the parameter scale and training time of each epoch for each method, combining with table 2, we found that when the performance are close, CQL with KAN-based actor has much smaller parameter scale that CQL with MLP-based actor. This is important when the occupy memory and computation amount is main concern to deploy learned policy (actor) on terminal device, such mobile phone. However, we should also notice that KAN-based CQL are more time consuming to train under current GPU settings.

\section{CONCLUSION}

Our study demonstrates that KAN can achieve performance close to the commonly used MLP with significantly fewer parameters in offline RL tasks. This provides us an option to choose the base networks according to the requirements of the tasks, when the performance is dominant, MLP-based methods are still the preferred ones, when the occupy memory and computation amount is the obstacle in deploying in the real world terminal devices, we can choose KAN-based actor without seriously affecting the performance.

KAN is originally designed for interpretability \cite{KAN_explainable}, which is not analyzed in this study. In the future, we can study how to build explainable offline RL methods with KAN, or learn by MLP-based offline RL methods and explain by KAN.

\bibliographystyle{plain}

\end{document}